\def\cvprPaperID{13}
\def\confName{CVPR}
\def\confYear{2023}

\def\paperTitle{Denoising diffusion models for out-of-distribution detection}
\def\authorBlock{
Mark S. Graham \\
King's College London\\
{\tt\small mark.graham@kcl.ac.uk}
\and
Walter H.L. Pinaya\\
King's College London\\
{\tt\small walter.diaz\_sanz@kcl.ac.uk}
\and
Petru-Daniel Tudosiu\\
King's College London\\
{\tt\small petru.tudosiu@kcl.ac.uk}
\and
Parashkev Nachev\\
University College London\\
{\tt\small p.nachev@ucl.ac.uk}
\and
Sebastien Ourselin\\
King's College London\\
{\tt\small sebastien.ourselin@kcl.ac.uk}
\and
M. Jorge Cardoso\\
King's College London\\
{\tt\small m.jorge.cardoso@kcl.ac.uk}
}

\newif\ifreview 
\newif\ifarxiv \newcommand{\arxiv}{\arxivtrue}
\newif\ifcamera 
\newif\ifrebuttal 

\arxiv%

\pdfoutput=1
\documentclass[10pt,twocolumn,letterpaper]{article}
\ifreview \usepackage[review]{cvpr} \fi
\ifarxiv \usepackage[pagenumbers]{cvpr} \fi
\ifrebuttal \usepackage[rebuttal]{cvpr} \fi
\ifcamera \usepackage{cvpr} \fi

\usepackage{graphicx}
\usepackage{amsmath}
\usepackage{amssymb}
\usepackage{booktabs}

\usepackage{times}
\usepackage{microtype}
\usepackage{epsfig}
\usepackage[table,xcdraw]{xcolor}
\usepackage{caption}
\usepackage{float}
\usepackage{placeins}
\usepackage{color, colortbl}
\usepackage{stfloats}
\usepackage{enumitem}
\usepackage{tabularx}
\usepackage{xstring}
\usepackage{multirow}
\usepackage{xspace}
\usepackage{url}
\usepackage{subcaption}
\usepackage{xcolor}
\usepackage[hang,flushmargin]{footmisc}

\ifarxiv  \fi

\newcommand{\R}[1]{{%
    \textbf{%
        \ifstrequal{#1}{1}{\textcolor{red}{R#1}}{%
        \ifstrequal{#1}{2}{\textcolor{blue}{R#1}}{%
        \ifstrequal{#1}{3}{\textcolor{magenta}{R#1}}{%
        \ifstrequal{#1}{4}{\textcolor{teal}{R#1}}{%
                           \textcolor{cyan}{R#1}%
        }}}}%
    }%
}}

\usepackage{xr-hyper}

\makeatletter
\newcommand*{\addFileDependency}[1]{
  \typeout{(#1)}
  \@addtofilelist{#1}
  \IfFileExists{#1}{}{\typeout{No file #1.}}
}

\makeatother

\usepackage[pagebackref,breaklinks,colorlinks]{hyperref}
\usepackage[capitalize]{cleveref}
\crefname{section}{Sec.}{Secs.}
\crefname{table}{Table}{Tables}
\crefname{figure}{Fig.}{Figs.}

\usepackage[page]{appendix} %

\begin{document}
\title{\paperTitle}
\author{\authorBlock}
\maketitle

\begin{abstract}
Out-of-distribution detection is crucial to the safe deployment of machine learning systems. Currently, unsupervised out-of-distribution detection is dominated by generative-based approaches that make use of estimates of the likelihood or other measurements from a generative model. Reconstruction-based methods offer an alternative approach, in which a measure of reconstruction error is used to determine if a sample is out-of-distribution. However, reconstruction-based approaches are less favoured, as they require careful tuning of the model's information bottleneck - such as the size of the latent dimension - to produce good results. In this work, we exploit the view of denoising diffusion probabilistic models (DDPM) as denoising autoencoders where the bottleneck is controlled externally, by means of the amount of noise applied. We propose to use DDPMs to reconstruct an input that has been noised to a range of noise levels, and use the resulting multi-dimensional reconstruction error to classify out-of-distribution inputs. We validate our approach both on standard computer-vision datasets and on higher dimension medical datasets. Our approach outperforms not only reconstruction-based methods, but also state-of-the-art generative-based approaches. Code is available at \url{https://github.com/marksgraham/ddpm-ood}.
\end{abstract}
\section{Introduction}
\label{sec:intro}
Out-of-distribution (OOD) detection plays a crucial role in the safe deployment of machine learning systems, ensuring that downstream models are only run on data sampled from the distribution they were trained on. OOD detection models can be broadly divided into unsupervised models, which only require in-distribution data for training, and supervised models, which require additional information such as classification labels or sample OOD data. Unsupervised models are appealing as they make no assumptions about the form OOD data will take or the type of downstream task (e.g. classification, segmentation) that will be performed.
\begin{figure}[tp]
    \centering
     \includegraphics[width=0.7\linewidth]{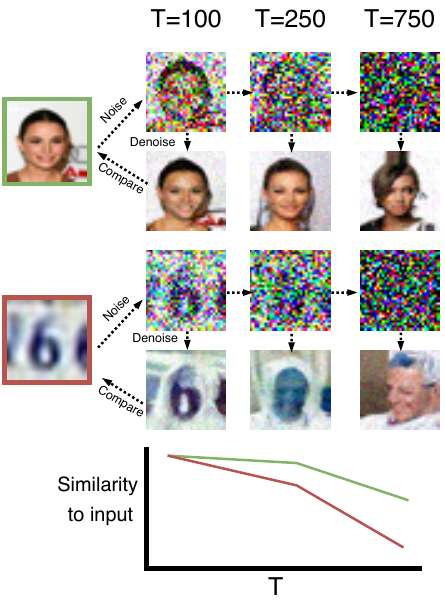}
    \caption{Reconstruction-based OOD detection, with the example of a model trained on CelebA. An in-distribution image from CelebA and an OOD image from SVHN are noised to various levels, reconstructed using the DDPM, and compared to the input. The similarity between inputs and reconstructions is plotted below.}
    \label{fig:abstract_figure}
\end{figure}

The current dominant approach in unsupervised OOD detection is the use of the likelihood or other metrics from a generative model trained on the in-distribution data. However, it has been shown these models can exhibit egregious failures, such as a model trained on CIFAR10 assigning higher likelihoods to samples from the SVHN dataset than samples from CIFAR10 itself \cite{nalisnick2018deep,choi2018waic, hendrycks2018deep}. A number of methods have been proposed to address these shortcomings \cite{ren2019likelihood,serra2019input,choi2018waic,nalisnick2018deep,morningstar2021density}. These models have shown better performance in empirical benchmarks, but recent theoretical work suggests that all these methods will remain vulnerable against at least some OOD data \cite{zhang2021understanding}.

Reconstruction-based methods offer an alternative approach to unsupervised OOD detection. They involve training a model to reconstruct in-distribution data and using the size of the reconstruction error to detect OOD inputs. However, compared to likelihood-based methods, reconstruction-based methods have received less attention in the literature \cite{yang2021generalized,salehi2021unified}. A likely reason is that these methods rely on an information bottleneck, such as a latent space that is smaller than the input, to effectively reconstruct in-distribution data but not OOD data. In practice, it is challenging to tune this bottleneck: too small and even the in-distribution data is poorly reconstructed, too large and even OOD data is successfully reconstructed. This need for tuning is undesirable and is likely a key reason these methods have typically been overlooked in favour of generative models for OOD detection.

Diffusion denoising probabilistic models (DDPM) \cite{sohl2015deep,ho2020denoising} present the information bottleneck issue in an interesting new light. These models are trained to incrementally remove noise from noised inputs. When an input is fully-noised, no information from the input itself is retained, and a DDPM will produce a new sample. However, when applying a DDPM to a partially-noised input, some information from the input is retained, and the denoising process is conditioned on that noisy input; that is, the model attempts to reconstruct the input. The amount of noise applied can be viewed as a variable information bottleneck. The bottleneck is not a property of the trained model itself, such as the size of the latent space in an autoencoder, but rather something we can control externally during model inference, meaning a single trained model can handle many different bottleneck levels. While this interpretation of DDPMs as autoencoders with an externally-controlled bottleneck has been previously discussed \cite{dieleman2022diffusion}, to our knowledge, no work has attempted to use this property to perform reconstruction-based OOD detection with DDPMs.

In this work, we apply DDPMs to perform reconstruction-based OOD detection. We measure the quality of a model's reconstructions of an input noised to a range of different levels and propose to use this set of reconstruction error metrics to determine whether an image is OOD.

\section{Related Work}
\label{sec:related}
Methods for OOD detection can be broadly categorised into supervised, requiring some additional labels or OOD data, or unsupervised, requiring only in-distribution data. This overview of related work focuses on unsupervised methods.

\subsection{Generative based}
A conceptually appealing approach to OOD detection involves fitting a generative model $p(\mathbf{x}; \theta)$ to a data distribution $\mathbf{x}$ and evaluating the likelihood of unseen samples under this model. The assumption is that OOD samples will be assigned a lower likelihood than in-distribution samples and can be identified using a simple threshold on this value \cite{bishop1994novelty}. It has since been demonstrated that OOD samples will not necessarily be assigned lower likelihoods than in-distribution data; for example, different families of generative models trained on CIFAR10 all assign higher likelihoods to images from SVHN  \cite{nalisnick2018deep,hendrycks2018deep,choi2018waic}. 

Subsequent studies have sought to address this shortcoming in generative models. One approach suggests the failures may be due to likelihood estimation errors and proposes using the Watanabe-Akaike Information Criterion (WAIC) across the likelihood estimates of an ensemble of models to identify OOD samples \cite{choi2018waic}. Other studies postulate that likelihood estimates are affected by population-level background statistics and propose using a likelihood ratio to remove this effect, either obtaining the denominator from an additional trained model \cite{ren2019likelihood} or a measure of image complexity \cite{serra2019input}. A closely related approach notes that these methods do not perform well for Variational Autoencoders (VAE) and seeks to develop a specific likelihood ratio for this class of model \cite{xiao2020likelihood}. Another related approach notes that lower-level model features dominate the likelihood and proposes the use of a hierarchical VAE and the requirement that a sample is in-distribution across all levels of the hierarchy \cite{havtorn2021hierarchical}. Another strand of work proposes flagging samples as OOD if their likelihoods do not lie within the typical set of a model - put simply, a sample may be considered OOD not only if its likelihood is lower than that of in-distribution data, but also if it is higher \cite{nalisnick2019detecting}. Follow-up work proposes assessing the typicality of multiple summary statistics from the model, not just the likelihood, to assess whether samples are OOD \cite{morningstar2021density}.

\subsection{Reconstruction based}
Reconstruction-based methods represent another paradigm for OOD detection. They involve training a model $R$ to reconstruct an input, $\hat{\mathbf{x}} = R(\mathbf{x})$. The intuition is that if $R$ contains an information bottleneck, such as a latent space of lower dimension than the input size, it will only be capable of faithfully reconstructing inputs from the distribution it was trained on, and will poorly reconstruct OOD inputs. This can be measured using some similarity measure between the input and reconstruction, $S(\mathbf{x},\hat{\mathbf{x}})$, and flagging inputs that show low similarity. 

Several works have highlighted practical issues in the use of reconstruction-based methods \cite{lyudchik2016outlier,pimentel2014review,denouden2018improving,zong2018deep}. If the information bottleneck is too small, a model cannot faithfully reconstruct even in-distribution samples. If the bottleneck is too large, the model is able to learn the identity function, allowing OOD samples to be reconstructed with low error. The result is that it is often necessary to perform a dataset-specific tuning process to produce a model that performs well, limiting the utility of such models in practice.

Some work has sought to address these issues. It has been suggested to use the Mahalanobis distance \cite{mahalanobis1936generalized} in an autoencoder's feature space as an OOD metric \cite{denouden2018improving}. Other work has sought to reduce the ability of an autoencoder to reconstruct OOD samples by introducing a memory module and forcing the decoder to decode directly from the memory, encouraging any reconstructions to look more similar to in-distribution data \cite{gong2019memorizing}.  However, none of these reconstruction-based works address the fundamental issue of bottleneck selection and require some tuning of the information bottleneck for the specific in-distribution/OOD dataset pairing being considered. 

Perhaps the work most closely related to ours is AnoDDPM \cite{wyatt2022anoddpm}. Whilst focusing on a slightly different problem of detecting localised anomalies, the work also proposes using reconstructions from a DDPM trained on in-distribution data. However, the authors propose to reconstruct from a single $t$-value, or noise bottleneck, with the choice of $t$ tuned to the dataset being considered. We seek to address this shortcoming in our work with DDPM-based OOD detection, which reconstructions from a range of noise values, obviating the need for any dataset-specific tuning. Concurrent to this work, a complementary approach that involves corrupting inputs and reconstructing with DDPMs was also developed \cite{liu2023unsupervised}.

\section{Method}
To enable reconstruction-based OOD detection that is not dependent on a fixed information bottleneck, we propose making use of a trained DDPM \cite{ho2020denoising} to reconstruct images. During training, samples $\mathbf{x_0}$ are degraded according to a fixed process with Gaussian noise according to a timestep $t$ and a noise variance schedule $\beta_t$ to produce noised samples $\mathbf{x}_t$, such that \begin{equation}
    q(\mathbf{x}_t|\mathbf{x}_0) = \mathcal{N}\left(\mathbf{x}_t| \sqrt{\bar{\alpha}_t}\mathbf{x}_0,(1-\bar{\alpha})\mathbf{I} \right)
\end{equation} 
where $0 \leq t \leq T$ and we define $\alpha_t :=1-\beta_t$ and $\bar{\alpha}_t := \prod_{s=1}^{t}\alpha_s$. The schedule $\beta_t$ is designed to increase with $t$ and have the property that the fully noised $\mathbf{x}_T$ is close to an isotropic Gaussian, $\mathbf{x}_T\sim\mathcal{N}\left(\mathbf{0}, \mathbf{I} \right)$; i.e. $\mathbf{x}_T$ contains no information about $\mathbf{x}_0$. We train a single network to iteratively reverse the diffusion process by estimating the parameters of the denoising step given $\mathbf{x}_t$ and $t$
\begin{equation}
p_\theta(\mathbf{x}_{t-1}|\mathbf{x}_{t})= \mathcal{N}\left(\mathbf{x}_{t-1}| \boldsymbol{\mu}_\theta(\mathbf{x}_t,t),\mathbf{\Sigma}_\theta(\boldsymbol{x}_t,t)\right)
\end{equation}

Given this trained model and a test input $\mathbf{x}_0$, we can sample a set of $\mathbf{x}_t$ for a range of values of $t$ and estimate their reconstructions, $\hat{x}_{0,t} = p_\theta(\mathbf{x}_0|\mathbf{x}_t)$. Measuring the similarity between each reconstruction and input $S(\hat{\mathbf{x}}_{0,t}, \mathbf{x}_t$) provides a range of similarity scores that can be used to decide whether $\mathbf{x}_0$ is in-distribution.

The advantage of such a method over other reconstruction-based methods is that the information bottleneck is no longer a property of the network itself, such as the latent-space dimension in an autoencoder, but an externally chosen factor, the amount of noise applied to the input. This allows for reconstructions from a wide number of information bottlenecks, obviating the need for the pre-selection of the appropriate bottleneck for a given dataset through the choice of model architecture.

\subsection{The diffusion model}
We use the DDPM model from \cite{rombach2022high}, which is a time-conditioned UNet \cite{ronneberger2015u}. While it is possible to optimise the variational bound on the negative log-likelihood, it has been found that a simplified training objective works well in practice, and we make use of it in this work. In this simplified scheme, the variance is fixed to time-dependent constants, $\mathbf{\Sigma}_\theta \left(\boldsymbol{x}_t,t)\right)= \dfrac{1-\bar{\alpha}_{t-1}}{1-\bar{\alpha}_t}\beta_t\mathbf{I}$ and the network directly predicts the added noise $\boldsymbol{\epsilon}$ at step $t$ \cite{ho2020denoising}:

\begin{equation}
    L_\text{simple}(\theta) = \mathbb{E}_{t,\mathbf{x}_0,\boldsymbol{\epsilon}}
    \left[
    \lVert\boldsymbol{
    \epsilon} -\boldsymbol{
    \epsilon}_{\theta}\left( \mathbf{x}_t\right)\rVert^2
    \right]
\end{equation}

\subsection{Multiple reconstructions}
Our method involves reconstructing an input $\mathbf{x}_t$ for multiple values of $t$. In the DDPM sampling scheme, $t$ steps are required to obtain the reconstruction $\hat{\mathbf{x}}_{0,t}$, with each step requiring an evaluation of the model. For a typical value $T=1000$ with reconstructions from 100 starting points equally spaced along the T-chain, $T_{\text{start}}=[10,20,30,...,990,1000]$ we would need 50500 model evaluations to obtain all the reconstructions; equivalent to the computation required to obtain 50.5 samples from the model. 

To expedite the process of obtaining multiple reconstructions, we make use of recent advances in fast sampling from diffusion models. In particular, we employ the PLMS sampler \cite{liu2021pseudo}, which has been shown to substantially reduce the number of sampling steps whilst maintaining or even improving sample quality. If we select $T=1000$ but choose to use just 100 sampling steps, we are still able to perform 100 reconstructions. However, these reconstructions can be done with just 5050 model evaluations, equivalent to the compute required to obtain 5.05 samples from the model with a DPPM sampler, a $10\times$ speed-up.

\subsection{Evaluating similarity}
It is common to use the mean-squared error (MSE) between the input and reconstruction to evaluate similarity. In this work, we also choose to use the LPIPS metric, which uses the distance between the deep features of a network (in this case, Alexnet \cite{krizhevsky2014one}) from two inputs as a measure of their perceptual similarity. LPIPS has been shown to correlate well with human evaluations of image similarity \cite{zhang2018unreasonable}. Using both MSE and LPIPS gives a total of 2$N$ similarity measurements per input for the $N$ reconstructions performed. We convert each measurement into a Z-score using the measurements from a validation set for each reconstruction and metric (MSE or LPIPS) separately. We average these $2N$ Z-scores to produce an OOD score for each input.

\section{Experiments and Results}
\label{sec:experiments_results}

\subsection{Experimental details}

\textbf{Datasets.} We evaluate our method both on a number of common computer vision benchmarks and, recognising that performance on these benchmarks don't necessarily reflect performance in the real world, on a set of higher dimension medical imaging datasets. For the computer vision benchmarks we use four in-distribution datasets: FashionMNIST \cite{xiao2017fashion}, CIFAR10 \cite{krizhevsky2009learning}, CelebA \cite{liu2015faceattributes}, and SVHN \cite{netzer2011reading}. For the grayscale FashionMNIST, we use MNIST \cite{lecun1998gradient} as an OOD dataset. For the other colour datasets, we use all other colour datasets as OOD datasets. CelebA images were resized to 32x32 to match the dimension of the other colour datasets.  We also use vertically- and horizontally-flipped versions of each in-distribution dataset as further OOD datasets, giving a total of 15 pairs of in vs out-of-distribution datasets. For the medical evaluation we used images from the MedNIST dataset, consisting of six classes of different organs and modalities: Hand X-ray, Abdomen CT, Chest X-ray, Chest CT, Breast MRI, and Head CT, with 10,000 images per class. We trained models on each class and evaluated against all other classes. We performed evaluations at the dataset's native dimension of $64\times64$, and also repeated the experiments on data upsampled to $128\times128$.

\textbf{Baselines.} We benchmark against both generative- and reconstruction-based approaches. We used the Glow architecture \cite{kingma2018glow} as the backbone for all generative approaches, following \cite{morningstar2021density}, and explored a number of methods in the literature to perform OOD using the trained model:
\begin{enumerate}
\item A threshold on the likelihood $p(\mathbf{x}; \theta)$ as a simple baseline \cite{bishop1994novelty}. 

\item The Watanabe-Akaike Information Criterion (WAIC), obtained by training an ensemble of 5 models and calculating the WAIC score as $\mathbb{E}_{\theta}[\log p(\mathbf{x}|\theta_n)] - \text{Var}_{\theta}[\log p(\mathbf{x}|\theta_n)]$ \cite{choi2018waic}. 

\item The single-sample typicality test where the typicality score for a sample $\mathbf{x}$ is given by $|\text{H}[p(\mathbf{x};\theta)] - \log p(\mathbf{x}; \theta)|$, where $\text{H}[p(\mathbf{x};\theta)]$ is calculated as an average over the training set \cite{nalisnick2018deep}.

\item Density-of-States Estimation (DoSE), to our knowledge the current state-of-the-art in unsupervised OOD detection. DoSE uses a density estimator to evaluate several features obtained from a model - for Glow models, the features are the model likelihood and its two constituent parts, the log-probability of the latent variable $Z$, and the log of the determinant between the input and $Z$. We used principal component analysis (PCA) to learn a whitening transform and trained a one-class support vector machine (SVM) on the transformed features from the validation set and used it to score new samples \cite{morningstar2021density}. As detailed in their paper, both the PCA and SVM used the default implementations in scikit-learn \cite{scikit-learn}.
\end{enumerate}

For the reconstruction-based approaches, we used:
\begin{enumerate}
    \item A threshold on the MSE from an autoencoder-based (AE) reconstruction  $||\mathbf{x}-\hat{\mathbf{x}}||_2$.
    
    \item A threshold on the Mahalanobis score, given by $\alpha D_M(E(\mathbf{x})) + \beta  ||\mathbf{x}-\hat{\mathbf{x}}||_2$ where $D_M(\mathbf{x})$ is the Mahlabonis distance from the latent space of $\mathbf{x}$ to the latent space of the training set and $\alpha$ and $\beta$ are constants set to the reciproal of the standard deviation of $D_M(E(\mathbf{x}))$ and $||\mathbf{x}-\hat{\mathbf{x}}||_2$ respectively, as evaluated on the validation set \cite{denouden2018improving}.
    
    \item The MSE from reconstructions using a memory-augmented autoencoder (MemAE), which augments the standard AE with a memory bank and seeks to decode samples from atomic elements in this bank \cite{gong2019memorizing}.
    
    \item The MSE from a reconstruction from $t=250$ for a DDPM, following AnoDDPM \cite{wyatt2022anoddpm}. As the method was intended for localised anomaly detection we modify it to make it more suitable for full-image OOD by training on Gaussian rather than simplex noise, as we found simplex noise hindered OOD performance. We refer to this as AnoDDPM-Mod.
\end{enumerate}

\textbf{Implementation details.} For our method, we used the DDPM model as described in \cite{rombach2022high}\footnote{\url{https://github.com/CompVis/latent-diffusion/}}. We used a 3-layer UNet with $[256,512,784]$ channels, with two residual blocks per layer and a single-headed attention block after each residual block in layers 2,3 of the downsampling branch and layer 1 of the upsampling branch. The timestep was sinusoidally embedded and passed through a two-layer MLP with a Swish activation function \cite{ramachandran2017searching} to create a 1024-dim embedding.  We used $T=1000$ during training and a linear noise schedule with $\beta_t$ varying between 0.0015 and 0.0195. All models were trained for 300 epochs using the Adam optimiser \cite{kingma2014adam} with a learning rate of $2.5e^{-5}$. At test time, we used the PLMS sampler with 100 timesteps and reconstructed from each of these 100 steps as starting points to produce 100 reconstructions per input. Our method was implemented in PyTorch and is available at \url{https://github.com/marksgraham/ddpm-ood}.

We implemented and trained all baselines to enable comparisons across the full set of dataset pairings considered, using the author's implementations when available. We trained Glow models\footnote{\url{https://github.com/y0ast/Glow-PyTorch}} using the architectural details outlined in \cite{nalisnick2019detecting,morningstar2021density}: we used 3 blocks (except for the FashionMNIST model, which used 2), each with 8 layers. Each layer used an activation norm, an inverted $1\times1$ convolution and affine coupling with 400 hidden channels per layer. Each model was trained for 100 epochs with a batch size of 64, using the Adamx optimiser with a learning rate of $5e^{-4}$, weight decay of $5e^{-5}$ and a 10 epoch warmup.

We trained\footnote{\url{https://github.com/donggong1/memae-anomaly-detection}} three-layer AEs with [128, 128, 256] features in the encoder and [128, 128, 3] in the decoder for colour datasets, and [32,16,8] in the encoder and [16,32,1] in the decoder for grayscale as in \cite{gong2019memorizing}. Each layer is followed by a batch normalisation \cite{ioffe2015batch} and a leaky ReLU activation, with upsampling implemented using transposed convolutions. The MemAE contained an additional memory module with sparse shrinkage. Following \cite{gong2019memorizing}, we used a memory size of 100 for the grayscale datasets and 500 for colour datasets, with a shrinkage threshold of $2.5e^{-3}$. Models were trained for 100 epochs with Adam with a learning rate of $1e^{-4}$ and a batch size of 10. 

\subsection{Results for computer vision datasets}

\begin{table*}
\centering
\small
 \setlength{\tabcolsep}{1pt}
\begin{tabular}{lcccccccccccccccc}
\toprule
& \multicolumn{3}{c}{FashionMNIST} &\multicolumn{4}{c}{CIFAR10} &\multicolumn{4}{c}{CelebA}&\multicolumn{4}{c}{SVHN} & Rank\\
\cmidrule(lr){2-4}\cmidrule(lr){5-8}\cmidrule(lr){9-12}\cmidrule(lr){13-16}

 & MNIST & VFlip & HFlip & SVHN& CelebA & VFlip & HFlip & CIFAR10& SVHN& VFlip& HFlip& CIFAR10& CelebA & VFlip& HFlip\\
\midrule
\textbf{Generative-based}\\
\midrule
Likelihood \cite{bishop1994novelty}& 8.5 &55.5 &51.1 &6.1 &52.3 &50.5 &50.0 &67.4 &5.9 &57.9 &50.1 &99.0 &\textbf{99.9} &50.2 &50.3 & 5.3\\
WAIC \cite{choi2018waic} &8.8 &55.4 &51.1 &6.0 &52.5 &50.4 &50.0 &67.0 &5.69 &57.8 &50.1 &99.0 &\textbf{99.9} &50.2 &50.3 & 5.7  \\
Typicality \cite{nalisnick2018deep} &81.1 &51.2 &49.6 &88.6 &40.3 &50.0 &50.0 &66.5 &88.8 &51.2 &50.0 &97.1 &99.8 &50.0 &49.9  & 6.6\\
Density-of-States \cite{morningstar2021density}&\textbf{98.1} &67.1 &55.3 &96.4 &54.6 &51.0 &50.1 &84.9 &99.4 &69.6 &49.9 &98.9 &\textbf{99.9} &50.3 &50.9 & 3.5\\

\midrule
\textbf{Reconstruction-based} \\
\midrule
AutoEncoder  & 75.0 &59.0 &50.4 & 3.2 &\textbf{75.3} &50.1 &49.9 & 42.1 &2.7 &53.2 &49.9 & \textbf{99.4} &\textbf{99.9} &49.9 &49.9 & 6.5\\
AutoEncoder Mahlabonis \cite{denouden2018improving} & 94.9 &79.5 &63.0 & 4.46 &71.8 &50.9 &50.0 & 64.2 &9.6 &71.6 &50.1 &99.3 &99.8 &50.6 &50.6 & 3.9 \\
MemAE \cite{gong2019memorizing}& 56.9 &59.0 &48.7 & 4.21 &69.4 &50.3 &49.9 & 51.5 &5.8 &56.4 &49.9 & 98.6 &99.5 &49.8 &49.7 & 7.3\\
AnoDDPM-Mod \cite{wyatt2022anoddpm} & 91.8 & 81.0 & 64.2 &37.8 & 60.2 & 54.2 & 50.5 & 80.2 & 67.3 & 78.1 & 49.4 & 90.4 & 94.2 & 50.2 & 52.7 & 4.2\\
DDPM (ours) &97.4 &\textbf{88.6} &\textbf{65.1} &\textbf{97.9} &68.5 &\textbf{63.2} &\textbf{50.5}&\textbf{99.0} &\textbf{100} &\textbf{93.3} &\textbf{50.3} &99.0 &99.6 &\textbf{58.2} &\textbf{61.6} & \textbf{1.9}\\
\midrule
\end{tabular}

\caption{
Results on computer vision datasets. AUC score for each comparison. Bold text indicates the highest value per column. The overall rank is calculated as the average across the ranks for each of the 15 comparisons (lower is better). 
}
\vspace{-0.05in}
\label{tab:main_results}
\end{table*}

Results are presented in  \cref{tab:main_results}, reported as AUC scores. The DDPM has the highest average rank across all experiments, with the state-of-the-art DoSE ranked second-highest. In a direct comparison, the DDPM outperforms DoSE on 13/15 dataset pairings. The increases in performance afforded by the DDPM are sometimes substantial, most notably when the OOD dataset is a vertically- or horizontally-flipped version of the in-distribution dataset. For example, on FashionMNIST vs VFlip, DDPM improves performance over DoSE from an AUC of 67.1 to 88.6 and on HFlip from 55.3 to 65.1. The DDPM is the only method able to perform substantially better than chance on certain pairings: for example, no other method scores higher than 50.9 on either SVHN vs VFlip or HFlip, whilst DDPM scores 58.2 and 61.6, respectively.

\begin{figure*}[tp]
    \centering
    \includegraphics[width=0.8\linewidth]{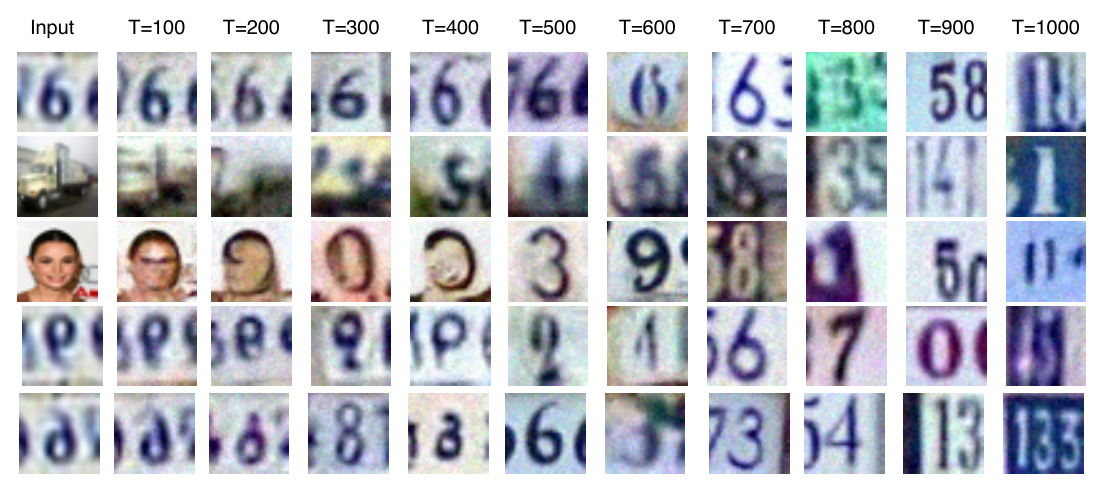}
    \caption{Example reconstructions from a model trained on SVHN for ten different $t$-values spaced equally across the chain. Plot shows an in-distribution input (top row) and OOD inputs from CIFAR10, CelebA, SVHN VFlip, and SVHN HFlip (rows 2-5).}
    \label{fig:results_demo_small}
\end{figure*}

\cref{fig:results_demo_small} shows some reconstructions from the model trained on the SVHN dataset. Reconstructions from all four models are included in Supplementary \cref{fig:results_demo_large}. Reconstructions of the in-distribution SVHN input still retain similarity to the input up until noising to $t=500-600$, whilst the OOD reconstructions start to look dissimilar to their inputs after noising to $t=100$ and bear almost no resemblance by $t=400$. The plot also shows that for $t \gtrapprox 700$ the noised images retain very little information from the input and the model outputs begin to resemble unconditioned samples more than they do reconstructions. This suggests that reconstructions from higher $t$ contribute little to the OOD signal and can potentially be discarded, though we view it as an advantage of our method that it performs well across dataset pairings without any post-hoc need for selecting the range of values to reconstruct from. We explore this further in \cref{performance}.

As widely reported \cite{nalisnick2018deep,choi2018waic,ren2019likelihood}, using the likelihood alone performs very poorly on a number of dataset pairings, such as FashionMNIST vs MNIST and CIFAR10 vs SVHN. Typicality substantially improves performance on the three dataset pairings that simple likelihood performs the worst on, but offers little improvements on other dataset pairings. In agreement with \cite{morningstar2021density} we found that WAIC provides little advantage over using just the likelihood, and sometimes degrades performance. 

All three AE-based reconstruction methods have inconsistent performance. They perform well on SVHN vs CIFAR10/CelebA but fail on other pairings, such as CIFAR10 vs SVHN. Inspection of the results reveals this class of method tends to perform better when the in-distribution datasets have simpler textures than the OOD datasets, as this causes higher OOD reconstruction error - this explains the good performance on SVHN vs CIFAR10/CelebA. Conversely, when in-distribution datasets are more complicated, the models can reconstruct simpler OOD datasets with low error and have poor OOD performance.

\subsection{Results on medical datasets}

\begin{table*}
\centering
\small
 \setlength{\tabcolsep}{1pt}
\begin{tabular}{lccccccccccccc}
\toprule
& \multicolumn{6}{c}{64 $\times$ 64} &\multicolumn{6}{c}{128 $\times$ 128} & Rank\\
\cmidrule(lr){2-7}\cmidrule(lr){8-13}

 & Abdomen& Breast& Chest & CXR& Hand & Head & Abdomen& Breast& Chest & CXR& Hand & Head\\
\midrule
\textbf{Generative-based}\\
\midrule
Likelihood \cite{bishop1994novelty}& 92.9 & 98.7  &97.1 &54.9 &73.6 &96.0 & 93.5 &99.5 &99.5 &90.1 &82.2 &93. & 5.8 \\
Typicality \cite{nalisnick2018deep} & 94.2 & 98.7  & 96.0 &70.3 &82.2 &90.1 & 93.9 &99.6 &99.5 &83.4 &75.9 &86.5  & 6.2\\
Density-of-States \cite{morningstar2021density}& \textbf{99.9} & 99.8 & 99.2 &76.5 &88.9 &\textbf{99.9} & 96.8 &99.9 &\textbf{99.7} &98.5 &79.5 &96.2 & 3.2\\

\midrule
\textbf{Reconstruction-based} \\
\midrule
AutoEncoder  & 99.0&99.4&95.6 & 96.9 & 89.5 & 90.0 & 99.2 & 99.7 & 97.8 & 94.7 & 91.5 & 90.4 & 4.8

\\
AutoEncoder Mahlabonis \cite{denouden2018improving} & 99.6 & 99.7 & 95.3 & 99.0 & 95.6 &93.6 & \textbf{99.9} & 99.8 & 99.5 & 99.2 & 95.1 & 94.1 & 3.3
\\
MemAE \cite{gong2019memorizing}& 75.8 & 99.7 & 85.5 &93.0 &87.8 &60.9 & 97.7 &98.0&96.0&82.2&85.7&58.1 & 6.6

\\
AnoDDPM-Mod \cite{wyatt2022anoddpm} & 96.4  & 88.4  & 96.7  & 99.0  & 98.5 & 87.7  & 98.5  & 98.9  & 95.0  & 99.4  & 99.1  & 94.1 & 4.6\\
DDPM (ours) & 99.7 & \textbf{100}  & \textbf{99.3}  & \textbf{100}  & \textbf{99.9} & 99.7 & 99.8 & \textbf{100} & 99.5 & \textbf{100} & \textbf{99.9} & \textbf{99.9} & \textbf{1.5}\\
\midrule
\end{tabular}

\caption{
Results on medical datasets, at both image size $64\times64$ and $128\times128$. Values for each dataset represent the AUC when considering all other datasets at the same image dimension as OOD datasets.}
\vspace{-0.05in}
\label{tab:mednist_results}
\end{table*}

Results are shown in \cref{tab:mednist_results}, with some sample reconstructions in Supplementary \cref{fig:results_demo_mednist}. We excluded WAIC from this comparison as \cref{tab:main_results} shows no benefit over simple likelihood whilst it requires training an ensemble of five models. The advantage of the DDPM over the other methods is more pronounced in these comparisons than in the computer vision datasets: it is the highest ranked method and also performs the most consistently, with 99.3 its lowest performance for any dataset pairing. A possible explanation for this is that at higher image sizes the noise level at which the model transitions from reconstructing to sampling is higher, so in-distribution data is successfully reconstructed in a greater number of the total reconstructions, resulting in a `cleaner' in-distribution signal. We consider it an advantage of the method that it performs well on $32\times32$ images where the reconstructions at higher noise values don't contribute to the OOD signal, whilst also being capable of using this information at higher image sizes - this shows the method does not require any dataset specific tuning.

\subsection{Bottleneck size}
To explore the importance of bottleneck size for reconstruction-based models, we considered the effect of changing the bottleneck on OOD performance. For the AE models we tried decreasing the size of the latent space, and both increasing and decreasing the memory size for MemAE models. For AnoDDPM-Mod we changed the $t$ value the image was noised to before reconstruction. Results are shown in Supplementary \cref{tab:memae_supplementary_results}. They show that for every method there was no single setting that worked well for all dataset pairings. For example, a 4-layer MemAE with a memory size of 2000 achieved 91.5 on FashionMNIST vs MNIST, substantially higher than the 3-layer model with recommended memory size, but this architecture's performance on SVHN vs CIFAR10 dropped to 89.3, making it the poorest-performing of any model (not just MemAE models) on this pairing. These results highlight the need for undesirable dataset-specific tuning for standard reconstruction-based OOD detection, and underscore the value of our method that automatically includes results from a range of noise bottlenecks, avoiding the need for fine-tuning.

\subsection{Performance and number of reconstructions}
\label{performance}
We evaluated the performance of the DDPM model as the number of reconstructions performed was reduced on the computer vision datasets. Results are shown in \cref{fig:performance_plot} with a full breakdown by dataset pairing included in Supplementary \cref{tab:model_evals_vs_performance}. Reconstruction starting points were uniformly sub-sampled across the range of possible starting points;  e.g. to reduce from 100 to 25 reconstructions we selected every fourth starting point. The results show that we can reduce the number of reconstructions from 100 to 25 with a mean drop in AUC across experiments of 0.58 (with AUC scores reported in the range of 0-100). Performing 25 reconstructions requires 1225 model evaluations, only slightly more than required to obtain a single sample from the model using the DDPM sampler over 1000 steps. Dropping even further to just 13 reconstructions gives a mean drop in AUC of 1.49, still outperforming the state-of-the-art DoSE method on 13/15 dataset pairings, but requiring just 637 evaluations, i.e. slightly over half those required to obtain a single sample from the model. Reducing the number of reconstructions below this point begins to affect performance substantially.
\begin{figure}[tp]
    \centering
    \includegraphics[width=0.85\linewidth]{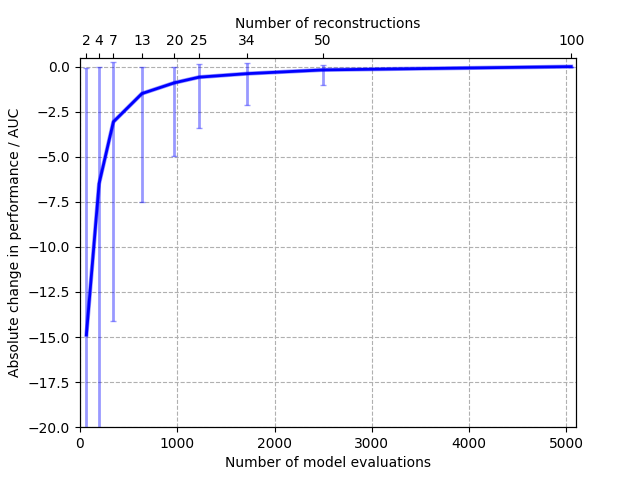}
    \caption{Results show the mean, minimum and maximum drop in absolute AUC as the number of reconstructions used to perform OOD detection is reduced, compared to results obtained using 100 reconstructions. The number of model evaluations is plotted on the lower x-axis, as this quantity directly correlates with total inference time. Full results are tabulated in the supplement.}
    \label{fig:performance_plot}
\end{figure}

\begin{figure}[tp]
    \centering
    \includegraphics[width=0.85\linewidth]{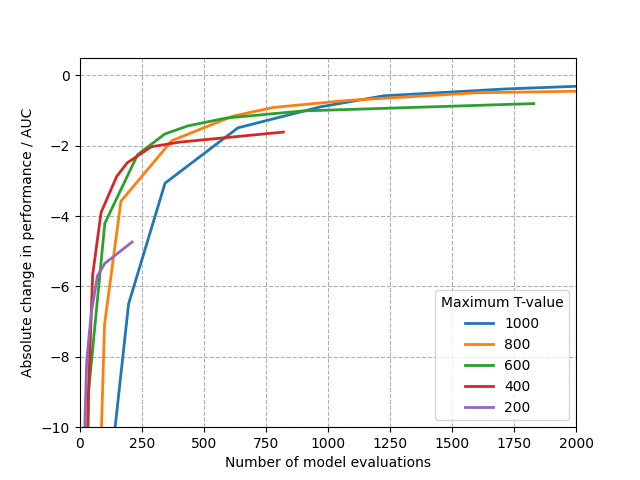}
    \caption{Results show the mean drop in absolute AUC, relative to the model with $\text{max}_T=1000$ and 100 reconstructions, as both $\text{max}_T$ and the number of reconstructions performed for a given $\text{max}_T$ are reduced. }
    \label{fig:performance_plot_maxt}
\end{figure}

We further explored reducing the maximum value of $t$ that reconstructions were performed from, $\text{max}_T$, motivated by the observation in \cref{fig:results_demo_small} that for the computer vision datasets the reconstructions from higher $t$-values do not seem heavily conditioned on the input and might not add to the OOD signal. \cref{fig:performance_plot_maxt} shows the mean drop in dataset performance, compared to the original model reported in \cref{tab:main_results}, as $\text{max}_T$ was reduced from 1000, and as the number of reconstructions was reduced for a given value of $\text{max}_T$. The results show that we can do better on the model-evaluations vs performance curve by reducing $\text{max}_T$, meaning for a limited compute budget it is better to focus on reconstructions from lower $t$-values. However, this result is likely dependent on the size of the input. There is a relationship between the amount of noise applied to an image and the size of features that are detectable in that image \cite{dieleman2022diffusion}, and our medical image experiments suggest that as input size increases, the maximum value of $t$ that provides useful OOD signal will also increase. Reducing $\text{max}_T$ did not lead to mean performance better than the model with $\text{max}_T=1000$ and 100 reconstructions, suggesting our approach is robust to the inclusion of non-informative signal and that there is no need for dataset-specific tuning of the bottlenecks. Detailed results are tabulated in Supplementary \cref{tab:model_evals_vs_performance}.

\subsection{Variants of the proposed method}

\begin{table*}
\centering
\small
 \setlength{\tabcolsep}{1pt}
\begin{tabular}{lccccccccccccccc}
\toprule
& \multicolumn{3}{c}{FashionMNIST} &\multicolumn{4}{c}{CIFAR10} &\multicolumn{4}{c}{CelebA}&\multicolumn{4}{c}{SVHN} \\
\cmidrule(lr){2-4}\cmidrule(lr){5-8}\cmidrule(lr){9-12}\cmidrule(lr){13-16}

 & MNIST & VFlip & HFlip & SVHN& CelebA & VFlip & HFlip & CIFAR10& SVHN& VFlip& HFlip& CIFAR10& CelebA & VFlip& HFlip\\
\midrule
\multicolumn{10}{l}{Metric: MSE+LPIPS}\\
\midrule

\multicolumn{1}{l}{Classification method:}\\
GMM &95.1 &82.3 &60.8 &96.8 &55.8 &53.1 &49.8 &98.3 &100 &88.9 &50.1 &97.6 &98.6 &52.3 &55.1 \\
SVM & 95.6 &82.5 &60.7 &96.4 &56.8 &52.8 &49.9 &97.5 &100 &87.6 &50.0 &97.8 &99.2 &52.6 &55.1 \\
Z-score average & 97.4 &88.6 &65.1 &97.9 &68.5 &63.2 &50.5 &99.0 &100 &93.3 &50.3 &99.0 &99.6 &58.2 &61.6 \\
\midrule
\multicolumn{10}{l}{Classification method: Z-score average}\\
\midrule
\multicolumn{1}{l}{Metric:}\\
MSE & 92.4 &83.6 &63.8 &40.2 &70.3 &57.2 &50.5 &86.3 &77.5 &86.5 &50.2 &97.3 &99.7 &58.8 &60.8 \\
LPIPS & 95.4 &81.5 &61.6 &98.5 &50.5 &58.6 &50.3 &98.3 &100 &87.6 &50.2 &92.3 &87.5 &53.4 &56.0 \\
MSE+LPIPS & 97.4 &88.6 &65.1 &97.9 &68.5 &63.2 &50.5 &99.0 &100 &93.3 &50.3 &99.0 &99.6 &58.2 &61.6 \\
\midrule

\end{tabular}

\caption{
AUC score for variants of the DDPM method. In the top set of results, the similarity metrics used are fixed to MSE+LPIPS, and we try three classification methods: a GMM, SVM, and a Z-score average. In the bottom set, the classification method is fixed to Z-score averaging, and we use three different similarity metrics: MSE, LPIPS, and MSE+LPIPS.
}
\vspace{-0.05in}
\label{tab:method_variants}
\end{table*}

\begin{figure}[tp]
    \centering
    \includegraphics[width=0.49\linewidth]{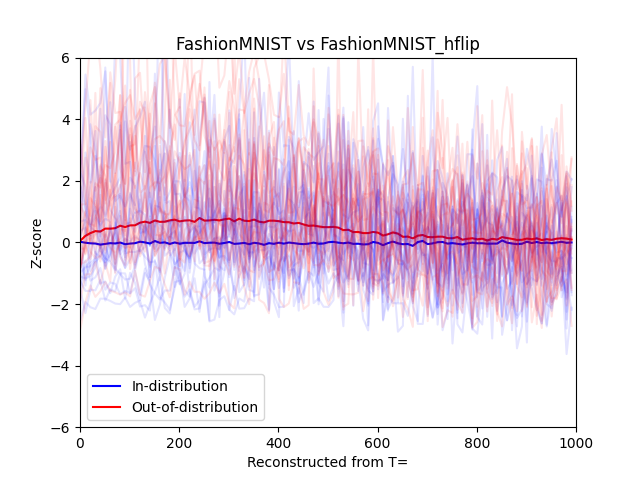}
    \includegraphics[width=0.49\linewidth]{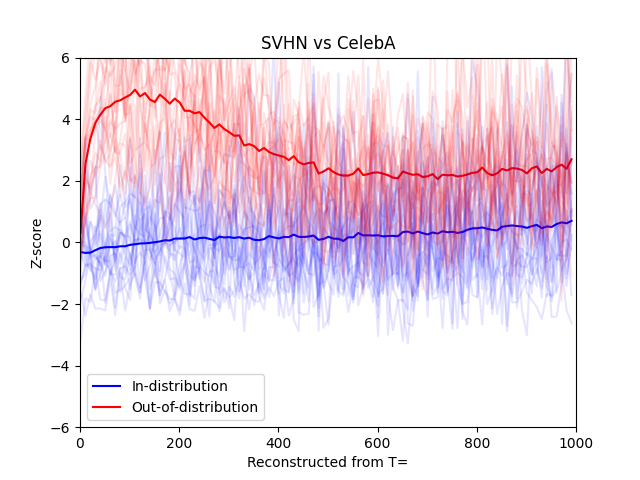}
    \caption{Plots of Z-scores of similarity for all 100 reconstructions of an image. Thin lines represent plots from individual inputs, thick lines show the means of these values. For FashionMNIST vs FashionMNIST HFlip (left), Z-score averaging performs much better than an SVM or GMM, whilst for SVHN vs CelebA (right), the difference in performance is small.}
    \label{fig:zscore_plots}
\end{figure}

We investigated the effects of changing the method used to classify OOD samples. We first tried using a one-class SVM or a Gaussian Mixture Model (GMM) to score samples. Results in \cref{tab:method_variants} show that, while all methods perform well, the GMM and SVM are consistently outperformed by the simple Z-score averaging method. \cref{fig:zscore_plots} shows the Z-scores for a dataset pairing where Z-score averaging substantially outperforms an SVM or GMM and a pairing where the difference is small. These plots suggest that Z-score averaging has an advantage when the signals from in-distribution and OOD inputs are noisy and overlap substantially, suggesting that the GMM/SVM overfit to these noisy signals, impairing classification performance. \cref{tab:method_variants} also shows that using MSE or LPIPS alone can be unreliable; both similarity measures have some dataset pairings they give poor performance on. Combining them is nearly always better than using either alone and provides robust performance above all dataset pairings; suggesting the two metrics are complementary.

\subsection{Limitations}
A key limitation of the method is that it is more computationally expensive than competing methods. This might limit the contexts for which the method can be used, even with the potential efficiency enhancements discussed in \cref{performance}. For example, the method may be suitable in medical imaging applications where it may be acceptable to flag an input as OOD within minutes, but less suitable in self-driving cars where predictions need to be made in real time. However, there is considerable scope for improving the efficiency of these models. The field of DDPMs is currently very active with substantial interest in fast sampling, and our method can directly benefit from advances in sampling speed \cite{kong2021fast,xiao2021tackling,zhang2022fast,salimans2021progressive,song2020denoising,vahdat2021score,dockhorn2022genie,watson2021learning,bao2021analytic}. Finally, while the steps along a single reconstruction chain cannot be parallelised, reconstructions from different starting points can be. This could reduce the reconstruction time down to the amount required to reconstruct the longest chain. 

\section{Conclusion}
\label{sec:conclusion}
In this work, we explored how DDPMs can be used to perform unsupervised OOD detection. We propose performing reconstructions of a number of inputs noised to different extents, addressing a drawback of standard reconstruction techniques that require the choice of a single, fixed bottleneck. We tested our method on both computer vision benchmarks and medical imaging datasets with a higher image size. These experiments show that our DDPM-based method outperforms not only reconstruction-based methods but also state-of-the-art generative approaches.

\noindent\textbf{Acknowledgements} MG, WS, PN, SO, and MJC are supported by a grant from the Wellcome Trust (WT213038/Z/18/Z).

{\small
\bibliographystyle{ieee_fullname}
\bibliography{11_references}
}

\ifarxiv \clearpage \appendix
\onecolumn
\FloatBarrier
\label{sec:appendix}
\section{Sample reconstructions on computer vision datasets.}
\begin{figure*}[h]
    \centering
    \includegraphics[width=0.8\linewidth]{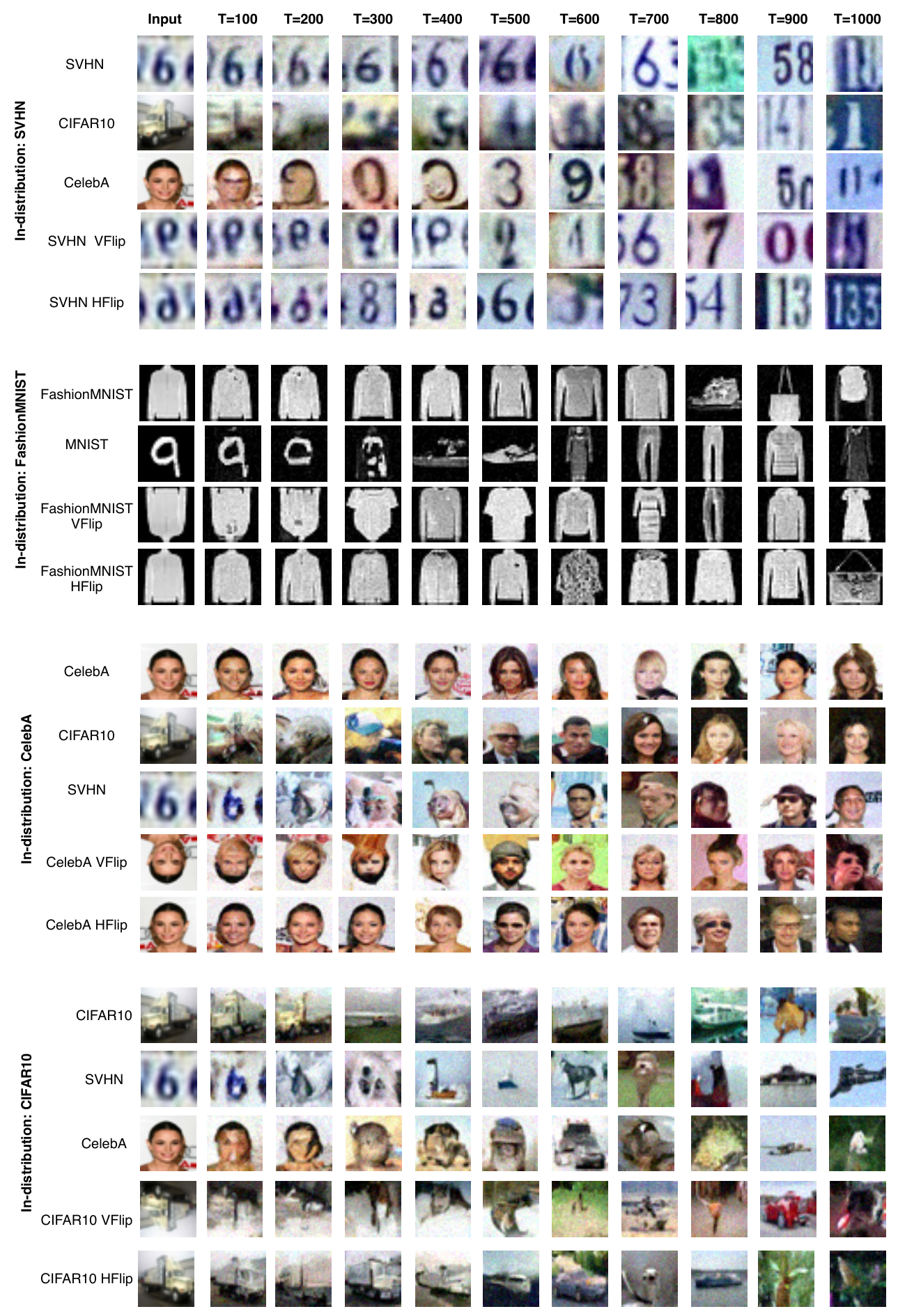}
    \caption{Example reconstructions from all models from ten different starting points spanning the full T-chain.}
    \label{fig:results_demo_large}
\end{figure*}

\FloatBarrier
\section{Sample reconstructions on medical datasets.}
\begin{figure*}[h]
    \centering
    \includegraphics[width=\linewidth]{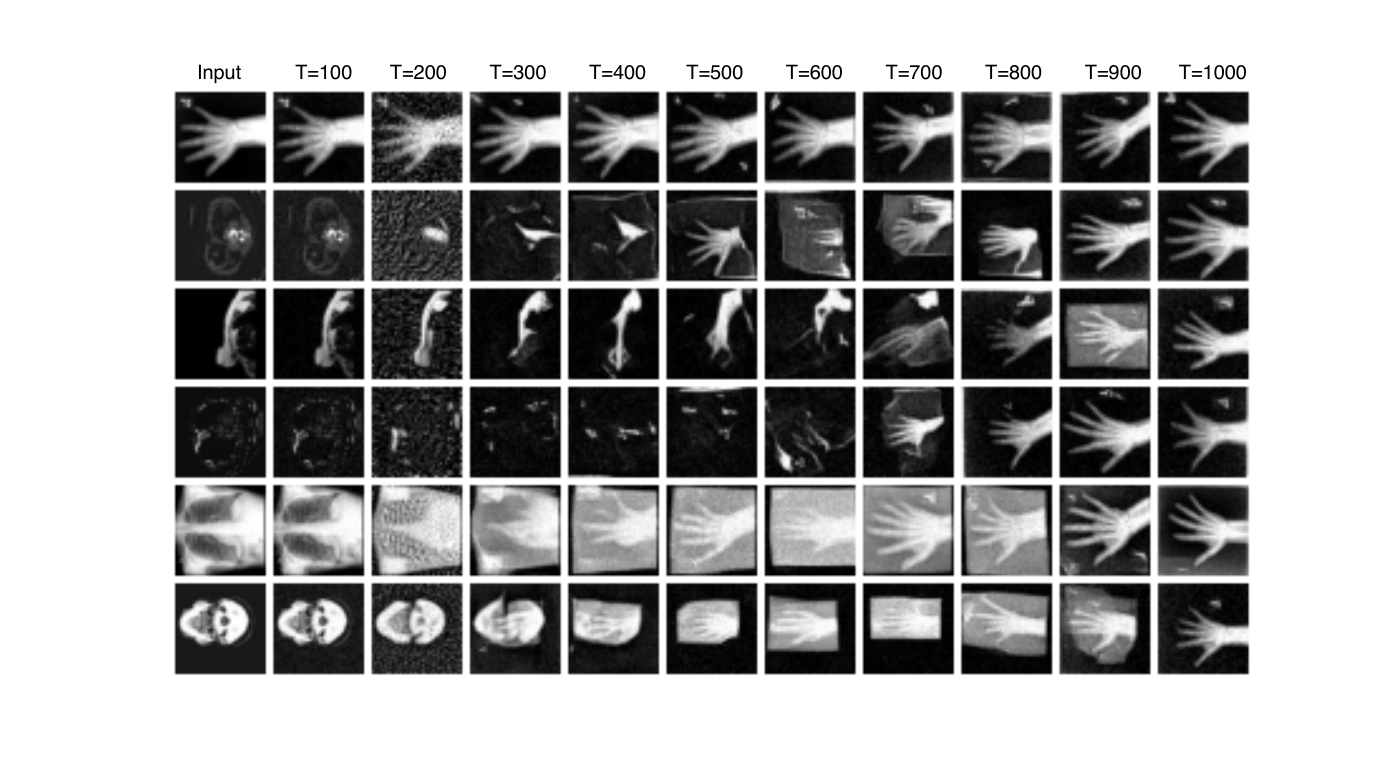}
    \caption{Example reconstructions from a model trained on MedNIST hand at $6x\times64$ for ten different $t$-values spaced equally across the chain. Plot shows an in-distribution input (top row) and OOD inputs from Abdomen CT, Breast MRI, Chest CT, Chest X-ray, Head CT (rows 2-6).}
    \label{fig:results_demo_mednist}
\end{figure*}

\newpage
\section{Effect of bottleneck size on the performance of reconstruction-based methods}
Full results are shown in \cref{tab:memae_supplementary_results}. For the AE the bottleneck is the size of the latent space; made smaller by increasing the number of layers in the AE. For the MemME we change both the size of the latent space and the size of the memory bank. For AnoDDPM-Mod we vary the amount of noise added to the image before reconstruction. Different bottleneck sizes could improve performance on some datasets, but at the expense of performance on others. For example, a 4-layer MemAE with a memory size of 2000 achieves 91.5 on FashionMNIST vs MNIST, substantially higher than the 3-layer model with recommended memory size, but this architecture's performance on SVHN vs CIFAR10 drops to 89.3, making it the poorest-performing of any model (not just MemAE models) on this pairing.  The results highlight that tuning the information bottleneck can improve performance on a given dataset, but at the expense of performance on other datasets.

\begin{table*}
\centering
\small
 \setlength{\tabcolsep}{1pt}
\begin{tabular}{lccccccccccccccc}
\toprule
& \multicolumn{3}{c}{FashionMNIST} &\multicolumn{4}{c}{CIFAR-10} &\multicolumn{4}{c}{CelebA}&\multicolumn{4}{c}{SVHN} \\
\cmidrule(lr){2-4}\cmidrule(lr){5-8}\cmidrule(lr){9-12}\cmidrule(lr){13-16}

 & MNIST & VFlip & HFlip & SVHN& CelebA & VFlip & HFlip & CIFAR10& SVHN& VFlip& HFlip& CIFAR10& CelebA & VFlip& HFlip\\

\midrule
AutoEncoder & 75.0 &59.0 &\textbf{50.4} & \textbf{3.2} &\textbf{75.3} &50.1 &\textbf{49.9} & 42.1 &2.7 &53.2 &\textbf{49.9} & \textbf{99.4} &\textbf{99.9} &49.9 &\textbf{49.9} \\
AutoEncoder  (4layer) & \textbf{95.1} &\textbf{71.9} &49.5 & 2.7 &66.4 &\textbf{50.4} &\textbf{49.9} & \textbf{65.2} &\textbf{6.7} &\textbf{68.5} &\textbf{49.9} & 98.8 &99.5 &\textbf{50.3} &\textbf{49.9} \\

\midrule
AutoEncoder Mahlabonis& 94.9 &\textbf{79.5} &63.0 & 4.5 &\textbf{71.8} &50.9 &\textbf{50.0} & 64.2 &9.6 &71.6 &\textbf{50.1} &\textbf{99.3} &\textbf{99.8} &50.6 &\textbf{50.6} \\
AutoEncoder Mahlabonis (4layer)& \textbf{98.3} &77.7 &\textbf{64.1} &\textbf{6.8} &69.0 &\textbf{51.4} &\textbf{50.0} & \textbf{73.3} &\textbf{19.5} &\textbf{78.7} &\textbf{50.1} & 96.9 &98.6 &\textbf{51.1} &\textbf{50.6} \\

\midrule
MemAE (mem size rec)& 56.9 &59.0 &48.7 & 4.21 &\textbf{69.4} &50.3 &49.9 & 51.5 &5.8 &56.4 &49.9 & 98.6 &99.5 &49.8 &49.7 \\
MemAE (mem size 2000)& 88.4 &58.2 &49.1 & 26.8 &61.2 &49.3 &50.0 & 67.2 &36.0 &63.1 &\textbf{50.0} & 95.2 &98.5 &50.1 &\textbf{50.7} \\
MemAE (mem size 50)&  28.6 &53.0 &49.9 & 7.2 &67.1 &\textbf{50.7} &49.8 &63.5 &33.3 &53.5 &49.8 & \textbf{98.7} &\textbf{99.6} &50.1 &50.0 \\
MemAE (4layer, mem size rec)&  77.3 &61.0 &51.1 & 7.1 &66.1 &50.6 &\textbf{50.1} & 75.1 &25.4 &73.3 &49.9 & 97.1 &98.6 &50.6 &50.2 \\
MemAE (4layer, mem size 2000)& \textbf{91.5} &\textbf{71.4} &\textbf{52.5} & \textbf{33.1} &66.4 &\textbf{50.7} &\textbf{50.1} & \textbf{78.6} &\textbf{55.8} &\textbf{80.3} &\textbf{50.0} & 89.3 &94.5 &\textbf{52.5} &50.4 \\
MemAE (4layer, mem size 50)&  87.6 &65.2 &51.3 & 7.0 &62.3 &50.6 &50.0 &71.7 &23.8 &64.4 &\textbf{50.0} & 97.2 &98.7 &50.3 &50.2 \\
\midrule
AnoDDPM-Mod $t=100$ & 76.6 & 79.5 & 62.2 & 23.6 & 55.8 & 53.2 & 50.3 & 79.2 & 52.7 & 77.3 & 49.7 & \textbf{95.3} & \textbf{98.1} & 46.8 & 48.1\\
AnoDDPM-Mod $t=250$ (rec) & \textbf{91.8} & \textbf{81.0} & \textbf{64.2} &37.8 & 60.2 & 54.2 & 50.5 & \textbf{80.2} & \textbf{67.3} & \textbf{78.1} & 49.4 & 90.4 & 94.2 & \textbf{50.2} & \textbf{52.7}\\
AnoDDPM-Mod $t=500$ & 81.5 & 68.8 & 58.4 & \textbf{51.0} & \textbf{64.1} & \textbf{54.6} & \textbf{50.8} & 71.8 & 68.4 & 72.6 & \textbf{50.9} & 74.8 & 79.9 & 50.0 & 51.6\\ 
\midrule

\end{tabular}

\caption{
These results show how performance on datasets vary as the information bottleneck is varied. Results show AUC scores. Bold text indicates highest value per column, done separately for each model class. Recommended memory size means 100 for FashionMNIST and 500 for other datasets, according to the settings in \cite{gong2019memorizing}.
}
\vspace{-0.05in}
\label{tab:memae_supplementary_results}
\end{table*}

\FloatBarrier

\section{Performance of DDPM vs number of reconstructions}
\FloatBarrier

\begin{table*}[t]
\centering
\small
 \setlength{\tabcolsep}{2pt}
\begin{tabular}{ccccccccccccccccc}
\toprule
Recons& Model evals& \multicolumn{3}{c}{FashionMNIST} &\multicolumn{4}{c}{CIFAR10} &\multicolumn{4}{c}{CelebA}&\multicolumn{4}{c}{SVHN} \\
\cmidrule(lr){3-5}\cmidrule(lr){6-9}\cmidrule(lr){10-13}\cmidrule(lr){14-17}

  && MNIST & VFlip & HFlip & SVHN& CelebA & VFlip & HFlip & CIFAR10& SVHN& VFlip& HFlip& CIFAR10& CelebA & VFlip& HFlip\\
\midrule
$\text{max}_{t}=1000$\\
100 & 5050 &97.4 &88.6 &65.1 &97.9 &68.5 &63.2 &50.5 &99.0 &100.0 &93.3 &50.3 &99.0 &99.6 &58.2 &61.6 \\
50 & 2500 &97.1 &88.1 &65.1 &97.8 &67.8 &63.0 &50.5 &98.9 &100.0 &92.9 &50.3 &99.0 &99.5 &58.0 &61.4 \\
34 & 1717 &97.0 &87.3 &64.5 &97.7 &67.3 &62.9 &50.7 &98.8 &100.0 &92.7 &50.5 &98.8 &99.4 &57.8 &60.9 \\
25 & 1225 &96.7 &86.9 &64.7 &97.6 &66.2 &62.3 &50.6 &98.7 &100.0 &92.2 &50.3 &98.7 &99.4 &57.9 &61.1 \\
20 & 970 &96.4 &86.2 &64.0 &97.3 &65.3 &62.0 &50.3 &98.6 &100.0 &91.8 &50.3 &98.6 &99.3 &57.7 &60.8 \\
13 & 637 &95.4 &84.6 &63.8 &97.2 &63.9 &61.0 &50.4 &98.3 &100.0 &90.7 &50.2 &98.2 &98.9 &57.0 &60.2 \\
7 & 343 &91.7 &80.6 &62.3 &96.6 &61.1 &59.4 &50.7 &97.3 &99.9 &87.4 &50.4 &96.6 &97.4 &55.9 &58.7 \\
4 & 196 &82.2 &73.5 &60.0 &95.2 &56.4 &56.4 &50.4 &94.7 &99.8 &80.9 &50.1 &92.0 &93.0 &54.0 &55.9 \\
2 & 66 &39.0 &60.7 &54.8 &91.8 &46.5 &52.6 &50.0 &87.8 &98.8 &69.7 &50.2 &83.3 &80.0 &51.4 &52.3 \\
\midrule
$\text{max}_{t}=800$\\
80 & 3240 &97.2 &88.6 &65.3 &96.7 &62.5 &62.1 &50.3 &98.7 &100.0 &92.7 &50.3 &99.2 &99.5 &60.2 &64.0 \\
40 & 1600 &96.9 &88.1 &65.3 &96.6 &62.0 &62.0 &50.4 &98.6 &100.0 &92.4 &50.3 &99.0 &99.4 &59.9 &63.7 \\
27 & 1080 &96.7 &87.5 &64.7 &96.5 &61.4 &61.8 &50.4 &98.5 &100.0 &92.1 &50.5 &98.9 &99.2 &59.7 &63.3 \\
20 & 780 &96.4 &87.1 &65.0 &96.3 &60.7 &61.4 &50.5 &98.4 &100.0 &91.7 &50.3 &98.8 &99.1 &59.6 &63.3 \\
16 & 616 &96.0 &86.5 &64.4 &96.1 &60.0 &61.3 &50.3 &98.3 &99.9 &91.3 &50.3 &98.7 &99.0 &59.5 &63.0 \\
10 & 370 &94.8 &85.1 &64.1 &95.7 &58.0 &60.2 &50.2 &97.9 &99.9 &90.0 &50.2 &98.2 &98.5 &58.9 &62.5 \\
5 & 165 &90.3 &81.7 &62.9 &94.7 &54.0 &58.7 &50.4 &96.4 &99.9 &86.6 &50.4 &96.6 &96.3 &58.1 &61.2 \\
3 & 99 &80.4 &74.9 &60.3 &93.6 &50.5 &56.2 &50.3 &93.7 &99.7 &80.3 &50.1 &92.2 &90.6 &55.5 &57.6 \\
2 & 66 &39.0 &60.7 &54.8 &91.8 &46.5 &52.6 &50.0 &87.8 &98.8 &69.7 &50.2 &83.3 &80.0 &51.4 &52.3 \\
\midrule
$\text{max}_{t}=600$\\
60 & 1830 &96.6 &88.4 &65.2 &95.0 &56.3 &61.4 &50.3 &98.2 &99.9 &92.1 &50.3 &99.0 &99.3 &62.0 &66.0 \\
30 & 900 &96.2 &88.0 &65.2 &94.8 &55.9 &61.2 &50.4 &98.1 &99.9 &91.7 &50.3 &98.9 &99.1 &61.6 &65.6 \\
20 & 590 &95.9 &87.5 &64.8 &94.7 &55.2 &61.1 &50.4 &97.9 &99.9 &91.3 &50.5 &98.7 &98.9 &61.5 &65.4 \\
15 & 435 &95.4 &87.0 &64.8 &94.6 &54.9 &60.8 &50.4 &97.8 &99.9 &90.9 &50.4 &98.6 &98.8 &61.2 &65.0 \\
12 & 342 &94.9 &86.5 &64.4 &94.4 &54.4 &60.9 &50.2 &97.6 &99.9 &90.3 &50.4 &98.4 &98.7 &61.2 &64.8 \\
8 & 232 &93.7 &85.2 &64.0 &94.3 &53.8 &60.0 &50.3 &97.3 &99.9 &89.4 &50.3 &98.0 &98.1 &60.1 &63.6 \\
4 & 100 &87.9 &81.7 &62.7 &93.2 &50.1 &58.6 &50.5 &95.6 &99.8 &85.4 &50.4 &96.2 &95.6 &59.0 &62.1 \\
2 & 34 &72.4 &75.0 &59.9 &90.5 &43.1 &56.0 &50.3 &91.0 &99.2 &76.6 &50.1 &90.3 &87.1 &56.7 &58.6 \\

\midrule
$\text{max}_{t}=400$\\
40 & 820 &93.2 &87.6 &64.3 &92.9 &50.5 &61.4 &50.3 &97.5 &99.9 &90.7 &50.4 &98.8 &99.2 &63.6 &67.5 \\
20 & 400 &92.3 &87.1 &64.1 &92.7 &50.1 &61.2 &50.4 &97.3 &99.9 &90.1 &50.4 &98.7 &99.0 &63.2 &66.9 \\
14 & 287 &92.3 &86.7 &63.8 &92.8 &50.1 &61.2 &50.3 &97.2 &99.9 &89.8 &50.5 &98.5 &98.8 &62.9 &66.7 \\
10 & 190 &90.2 &86.0 &63.6 &92.4 &49.1 &60.8 &50.5 &97.0 &99.8 &89.0 &50.5 &98.4 &98.7 &62.7 &66.2 \\
8 & 148 &88.8 &85.3 &63.3 &92.3 &48.6 &60.7 &50.1 &96.7 &99.8 &88.3 &50.2 &98.2 &98.4 &62.5 &65.8 \\
5 & 85 &84.4 &83.7 &62.6 &91.7 &47.0 &59.9 &50.4 &96.0 &99.7 &86.4 &50.3 &97.6 &97.6 &61.7 &64.7 \\
3 & 51 &79.8 &80.6 &61.6 &91.3 &45.5 &58.4 &50.3 &94.4 &99.6 &82.7 &50.2 &95.7 &94.7 &59.8 &62.5 \\
2 & 34 &72.4 &75.0 &59.9 &90.5 &43.1 &56.0 &50.3 &91.0 &99.2 &76.6 &50.1 &90.3 &87.1 &56.7 &58.6 \\

\midrule
$\text{max}_{t}=200$\\
20 & 210 &69.7 &82.8 &61.7 &90.8 &45.6 &60.5 &50.3 &96.2 &99.5 &87.4 &50.3 &98.4 &99.1 &62.7 &65.7 \\
10 & 100 &66.5 &81.9 &61.4 &90.5 &45.0 &60.2 &50.3 &95.8 &99.5 &86.4 &50.3 &98.1 &98.9 &62.1 &64.9 \\
7 & 70 &64.8 &81.3 &61.1 &90.5 &44.7 &60.1 &50.4 &95.6 &99.5 &85.6 &50.3 &97.8 &98.5 &61.7 &64.6 \\
5 & 45 &58.0 &79.7 &60.5 &90.1 &43.4 &59.2 &50.3 &95.0 &99.3 &84.1 &50.5 &97.5 &98.1 &61.1 &63.5 \\
4 & 34 &53.6 &78.5 &60.3 &90.0 &42.6 &59.0 &50.0 &94.5 &99.2 &82.8 &50.1 &97.0 &97.5 &60.3 &62.6 \\
3 & 27 &53.0 &77.7 &59.7 &89.7 &42.0 &58.3 &50.2 &93.8 &99.2 &81.1 &50.4 &96.3 &96.2 &60.0 &62.4 \\
2 & 18 &46.3 &74.6 &58.4 &89.0 &40.2 &57.0 &50.1 &91.6 &99.0 &76.5 &50.3 &93.8 &91.9 &58.5 &60.7 \\

\midrule
\end{tabular}

\caption{
Variation in the performance of the DDPM OOD detection as the number of reconstructions used is changed. Values reported for each dataset pairing ar AUCs. The first row ($\text{max}_{t}=1000$, 100 reconstructions) are the parameters for the results reported in \cref{tab:main_results}.
}
\vspace{-0.05in}
\label{tab:model_evals_vs_performance}
\end{table*}

Results in \cref{tab:model_evals_vs_performance}, both for reducing the number of reconstructions performed and for reducing the maximum value of $t$ that reconstructions are performed from, $\text{max}_T$. The results demonstrate the number of reconstructions/model evaluations can be reduced substantially with limited effect on OOD performance.
\FloatBarrier
 \fi

\end{document}